\def\year{2020}\relax
\documentclass[letterpaper]{article} 
\usepackage{aaai20}  
\usepackage{times}  
\usepackage{helvet} 
\usepackage{courier}  
\usepackage[hyphens]{url}  
\usepackage{graphicx} 
\usepackage{graphicx}  
\usepackage{multirow}
\usepackage{amsmath}
\title{Synchronous Speech Recognition and Speech-to-Text Translation \\ with Interactive Decoding}
\author{
	\Large \textbf{
		Yuchen Liu,\textsuperscript{\rm 1,2}\thanks{This work is done while Yuchen Liu is doing research intern at Baidu Inc.} 
		Jiajun Zhang,\textsuperscript{\rm 1,2} 
		Hao Xiong,\textsuperscript{\rm 4} 
		Long Zhou,\textsuperscript{\rm 1,2}} \\ 
	\Large \textbf{
		Zhongjun He,\textsuperscript{\rm 4} 
		Hua Wu,\textsuperscript{\rm 4} 
		Haifeng Wang,\textsuperscript{\rm 4} 
		and Chengqing Zong\textsuperscript{\rm 1,2,3}}\\ 
	\textsuperscript{\rm 1} National Laboratory of Pattern Recognition, Institute of Automation, CAS\\
	\textsuperscript{\rm 2} University of Chinese Academy of Sciences\\ 
	\textsuperscript{\rm 3} CAS Center for Excellence in Brain Science and Intelligence Technology\\
	\textsuperscript{\rm 4} Baidu Inc., No. 10, Shangdi 10th Street, Beijing, China\\
	\{yuchen.liu, jjzhang, long.zhou, cqzong\}@nlpr.ia.ac.cn\\
	\{xionghao05, hezhongjun, wu\_hua, wanghaifeng\}@baidu.com
}
 
\begin{document}

\maketitle

\begin{abstract}
Speech-to-text translation (ST), which translates source language speech into target language text, has attracted intensive attention in recent years. 
Compared to the traditional pipeline system, the end-to-end ST model has potential benefits of lower latency, smaller model size, and less error propagation. 
However, it is notoriously difficult to implement such a model without transcriptions as intermediate.
Existing works generally apply multi-task learning to improve translation quality by jointly training end-to-end ST along with automatic speech recognition (ASR). However, different tasks in this method cannot utilize information from each other, which limits the improvement. Other works propose a two-stage model where the second model can use the hidden state from the first one, but its cascade manner greatly affects the efficiency of training and inference process.
In this paper, we propose a novel interactive attention mechanism which enables ASR and ST to perform synchronously and interactively in a single model.
Specifically, the generation of transcriptions and translations not only relies on its previous outputs but also the outputs predicted in the other task. Experiments on TED speech translation corpora have shown that our proposed model can outperform strong baselines on the quality of speech translation and achieve better speech recognition performances as well.
\end{abstract}

\section{Introduction}
\label{introduction}
Speech-to-text translation (hereinafter referred to as speech translation)
aims to translate a speech in source language into a text in target language, which can help people efficiently communicate with each other in different languages.
The traditional approach is a pipeline system composed of an automatic speech recognition (ASR) model and a text machine translation (MT) model. 
In this approach, two models are independently trained and tuned, leading to the problem of time delay, parameter redundancy, and error propagation.
In contrast, end-to-end ST model has potential advantages to alleviate these problems. Recent works have  emerged rapidly and shown promising performances 
\cite{weiss2017sequence,berard2018end,bansal2018pre,jia2019leveraging,sperber2019attention}.

\begin{figure}[t]
	\centering
	\includegraphics[width=\columnwidth]{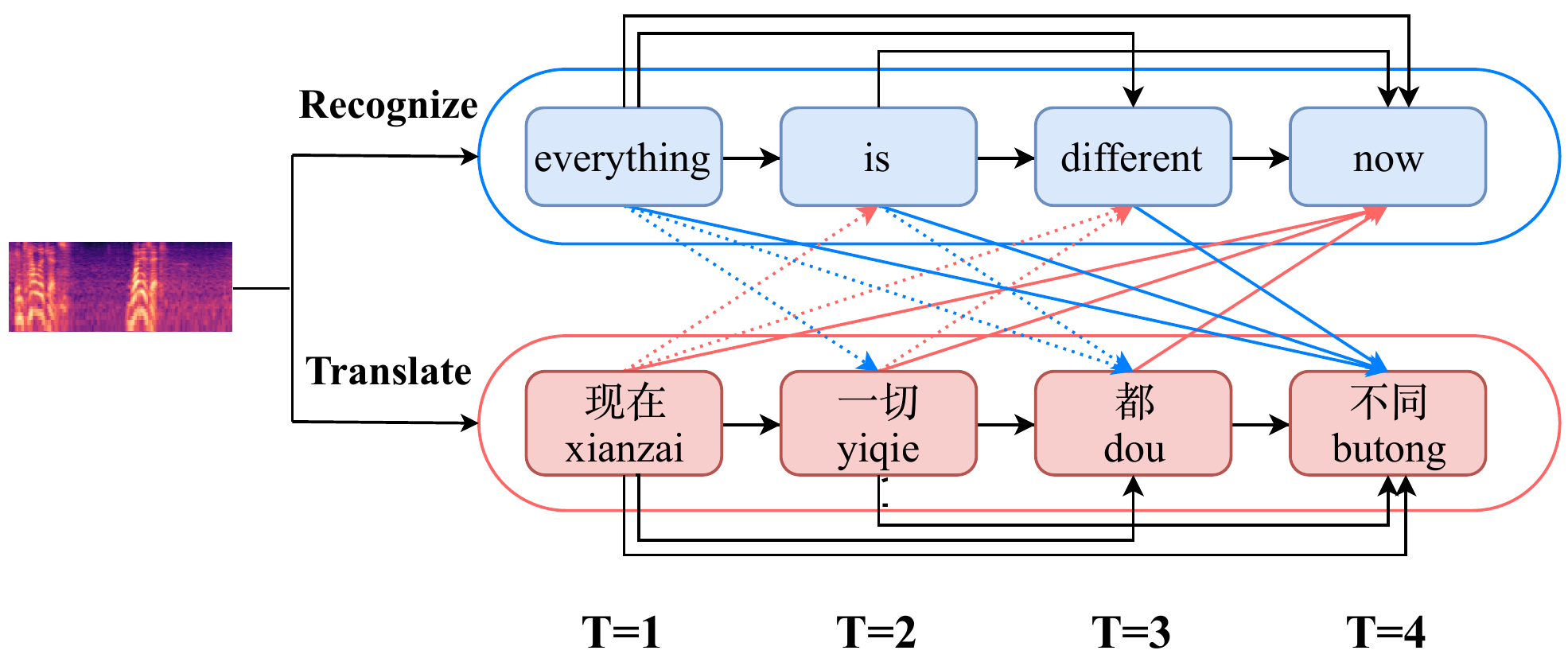}
	\caption{An example of an English speech recognition and the corresponding translation in Chinese, where the outputs in two tasks can interact with each other.}
	\label{fig:example}
\end{figure}

Despite the advantages, it is notoriously difficult to implement an end-to-end ST model which does not use transcriptions as intermediate, and its performance is generally limited. 
Previous studies resort to pretraining or multi-task learning to improve the translation quality. They either apply a pretrained encoder trained on ASR data \cite{bansal2018pre}, or jointly train with ASR to obtain a better acoustic model, or with MT to acquire a better language model \cite{weiss2017sequence,anastasopoulos2018leveraging,berard2018end}. However, the basic unit shared between different tasks is module parameters. Different tasks in this method cannot utilize information from each other. To alleviate this flaw, several studies propose analogous two-stage model  \cite{kano2017structured,anastasopoulos2018tied,sperber2019attention}.
In this model, decoder in the first stage performs recognition and generates a hidden state with which the second decoder conducts translation. Although the translation quality can be improved with additional information from the first decoder, the second decoder needs to wait until the complete transcription is recognized, greatly limiting the efficiency of training and inference process. In addition, this model can only make the translation process utilize information from recognition process, but cannot for the other direction.

However, we find that the generation process in ASR and ST can help each other: (1) the generation of speech translation would become easier with additional information from the transcribed words than just from the speech signal, (2) the translated words can also assist the recognition process. As the example shown in Figure \ref{fig:example}, the input is a complete speech utterance in English and the outputs in two tasks can interact with each other. When translating Chinese word ``yiqie" (the meaning of everything) at step $T=2$ in the ST task, the already transcribed word ``everything" at step $T=1$ in the transcription can provide the additional context. For ASR task, the translated word ``xianzai" at step $T=1$ can also help to recognize ``now" at step $T=4$. Therefore, if the generation of two tasks can interact with each other, the quality of transcription and translation can both be improved.

To this end, we propose a novel interactive learning model which can perform speech recognition and speech translation synchronously and interactively. Compared with the traditional multi-task learning model which shares part of parameters and treats different tasks separately, tasks in our approach can exchange the information of each other. With an interactive attention sub-layer, translation decoder in our model predicts next word with the transcribed words as auxiliary information, and for recognition decoder vice versa. Therefore, at each step, word prediction in  each task not only relies on its previously generated outputs, but also the outputs in the other task. 
Furthermore, we introduce a wait-$k$ policy where the generation process of speech translation is always $k$ steps later than speech recognition, so that the translation decoder can attend to more transcribed words. 
We conduct extensive experiments to verify the effectiveness of our proposed approaches on new TED English-to-German/French/Chinese/Japanese speech translation corpora. 

Our main contributions are summarized as follows:
\begin{itemize}
	\item We propose an interactive learning model which can conduct speech recognition and speech translation interactively, enhancing the quality of both tasks. 
	\item Different from traditional multi-task learning model which generates transcriptions or translations separately, our method can simultaneously generate both transcriptions and translations in one model. 
	\item Experiments on four language pairs have demonstrated that our model can outperform strong baselines, including the pipeline system, the pretrained end-to-end ST model, the traditional multi-task learning model, and the two-stage model.
\end{itemize}

\section{Related Work}
\subsubsection{Speech Translation} Speech translation has traditionally been approached through a pipeline system which consists of an ASR model and a text MT model \cite{sutskever2014sequence,Bahdanau:2015,chan2016listen,chiu2017state,vaswani2017attention}. 
Recent works have shown the feasibility of collapsing the cascade system into an end-to-end model. 
The first conjecture was proposed by \citeauthor{1999The}~\shortcite{1999The} who presumed that end-to-end speech translation is possible to implement with the development of memory, computation speed, and representation methods.
It is not until 2016 that~\citeauthor{berard2016listen}~\shortcite{berard2016listen} realized the first pure end-to-end model without using any source transcriptions.
Considering its notorious difficulty, the performance of end-to-end ST model is generally limited. Several work proposed a variety of approaches to improve the translation quality. Some applied multi-task learning to train speech translation jointly with ASR \cite{weiss2017sequence,anastasopoulos2018leveraging,berard2018end}. Others attempted to pretrain ST model with extra ASR data to promote acoustic model, or with target sentences to improve language model \cite{bansal2018pre,jia2019leveraging}. \citeauthor{liu2019end}~\shortcite{liu2019end} proposed to use a text MT model as teacher model to instruct ST model through knowledge distillation.

An intuition is that speech translation can become easier if the model has access to the transcription as intermediate. Therefore, several researchers proposed two-stage models where the first decoder is used to recognize transcriptions and the second decoder conducts translating with the hidden state in the former stage. 
\citeauthor{kano2017structured}~\shortcite{kano2017structured} first proposed the basic two-stage model and used pretraining strategy for the individual sub-models.  \citeauthor{anastasopoulos2018tied}~\shortcite{anastasopoulos2018tied,anastasopoulos2018leveraging} employed a triangle model on low-resource speech translation. \citeauthor{sperber2019attention}~\shortcite{sperber2019attention} further applied an attention-passing mechanism which can integrate auxiliary data and improve model robustness.
However, the second decoder needs to wait until the complete transcription is recognized, which greatly affects the training and inference efficiency. Besides, it can only utilize transcriptions to improve translation quality but leaves the recognition task alone. As shown in Figure \ref{fig:example}, the outputs of recognition and translation are complementary and can benefit each other. Therefore, it is reasonable to improve the quality of both tasks through interactive learning.

\subsubsection{Synchronous Inference}
\citeauthor{zhou2019synchronous}~\shortcite{zhou2019synchronous} proposed a synchronous bidirectional inference model in which left-to-right and right-to-left inferences perform in parallel.
The two decoding directions can help each other, and make full use of the target-side history and future information during translation. 
\citeauthor{zhang2019synchronous}~\shortcite{zhang2019synchronous} further applied this inference model on other sequence generation tasks, such as summarization, obtaining significant improvement as well. However, their works are conducted on the same task with outputs in different directions. 
The most related work with us is from  \citeauthor{wang2019synchronously}~\shortcite{wang2019synchronously} who synchronously performed multilingual translation within a beam.
In our work, we have two different tasks and aim to implement speech recognition and speech translation in one model synchronously.

\begin{figure*}[t]
	\centering
	\includegraphics[width=1.95\columnwidth]{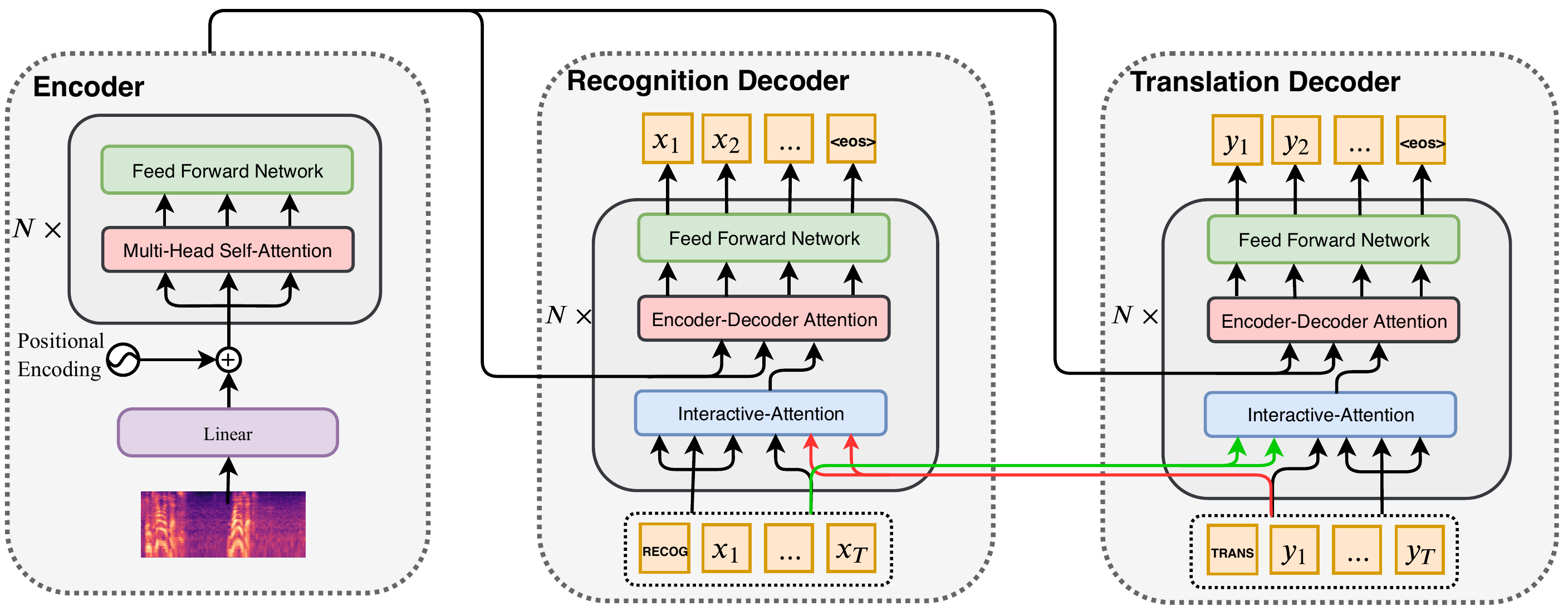}
	\caption{The model architecture of our method which is adopted from Transformer. The left part is the encoder which takes speech features as input and is shared by speech recognition model and speech translation model. The middle and right parts are two decoders where the middle is recognition decoder and the right is the translation decoder. There is an interactive attention sub-layer between two decoders which makes the decoders can utilize information from each other.}
	\label{fig:transformer}
\end{figure*}

\section{Background}
\label{sec:transformer}
Considering that Transformer model is now the state-of-art model in MT field \cite{vaswani2017attention}, and also shows a superior performance in ASR filed  \cite{dong2018speech,zhou2018syllable}, we adopt Transformer model as the core structure.  However, our proposed approach can be applied to any encoder-decoder architectures. 

The Transformer follows the typical encoder-decoder architecture. The encoder first maps the input sequence $\mathbf{\mathit{I}}=(i_1,i_2,\cdots,i_n)$ into a sequence of continuous representations $\mathbf{\mathit{Z}}=(z_1,z_2,\cdots,z_n)$, from which  the decoder generates the output sequence $\mathbf{\mathit{O}}=(o_1,o_2,\cdots,o_m)$ one word at a time. In Transformer, the encoder includes $N$ layers and each layer is composed of two sub-layers: the self-attention sub-layer and the feed-forward sub-layer.
The decoder also consists of $N$ layers and each layer has three sub-layers. The first one is the masked self-attention sub-layer, which adds masks to prevent present positions from attending to the future positions during training. The second is the encoder-decoder attention sub-layer, followed by the feed-forward sub-layer. Residual connection and layer normalization are employed around each sub-layer in the encoder and decoder.

The calculation process of three attention sub-layers can be formalized into the same formula as, 
\begin{equation}
\mathrm{Attention(\mathbf{Q},\mathbf{K},\mathbf{V})}=\mathrm{softmax}(\frac{\mathbf{Q}\mathbf{K}^T}{\sqrt{d_k}})\mathbf{V} 
\label{equ:Attention}
\end{equation}
where $\mathbf{Q}$, $\mathbf{K}$ and $\mathbf{V}$ denotes the query, key, and value respectively. $d_k$ is the dimension of the key.
The feed-forward sub-layer is then applied to yield the output of a whole layer. And softmax function is employed to predict the final output.

It is worth noting that for the self-attention sub-layer, the query, key, and value are the hidden representation from the same layer. For encoder-decoder sub-layer, the query is the hidden representation from the masked self-attention sub-layer in the decoder, the key and value are from the top layer in the encoder. 

\section{Our Approach}
In this section, we propose a novel framework to implement interactive learning for speech recognition and speech translation during training and inference, which is shown in Figure~\ref{fig:transformer}. Before we introduce this framework in detail, we first introduce how Transformer model is applied to the ASR, MT, and ST tasks.

\subsection{ASR, MT, and ST Task}
Speech recognition, text machine translation, and speech translation tasks can all adopt Transformer model, while different tasks have different input sequences $\mathbf{\mathit{I}}$ and output sequences $\mathbf{\mathit{O}}$. Specifically,
\begin{itemize}
	\item For ASR task, the input sequence $\mathbf{\mathit{I}}=\mathbf{\mathit{S}}=[s_1,\dots,s_T]$ is a sequence of speech features, where $T$ is the frame number of speech sequence. Specifically, the speech feature is first converted from raw speech signal by applying log-Mel filterbanks with mean and variance normalization. Frame stack and downsampling are used  to reduce the input length similar with~\citeauthor{sak2015fast}~\shortcite{sak2015fast}, resulting in a sequence with dimension of $d_{\textrm{filterbank}} \times \textrm{num}_{\textrm{stack}}$. 
	The output sequence $\mathbf{\mathit{O}}=\mathbf{\mathit{X}}=[x_1,\dots,x_N]$ is the corresponding transcription, where $N$ is the source sentence length.
	\item For  MT task, the input sequence $\mathbf{\mathit{I}}=\mathbf{\mathit{X}}=[x_1,\dots,x_N]$ is  the transcription in source language and the output sequence $\mathbf{\mathit{O}}=\mathbf{\mathit{Y}}=[y_1,\dots,y_M]$ is the corresponding translation in target language, where $M$ is the target sentence length.
	\item For end-to-end ST task, the input sequence $\mathbf{\mathit{I}}=\mathbf{\mathit{S}}=[s_1,\dots,s_T]$ is the same with ASR task and the output sequence $\mathbf{\mathit{O}}=\mathbf{\mathit{Y}}=[y_1,\dots,y_M]$ is the corresponding translation in target language.
\end{itemize}

In addition to the end-to-end model, ST task can also be implemented in a pipeline approach, where the speech utterance is first transcribed by an ASR model and then passed to a MT model. Another method is the multi-task learning model where the ASR model and ST model are combined with a shared encoder and trained jointly. 

\subsection{Interactive Learning Model}
In the traditional multi-task learning, different tasks are trained independently with shared parameters. However, as discussed in Section~\ref{introduction}, the output in one task is complementary with that in the other which can assist the prediction. Therefore, it is reasonable to improve the performances of both tasks by interactively exchanging information from each other. Besides, the traditional multi-task learning can only perform one task during inference, while sometimes the transcription and translation are required at the same time. To solve these problems, we propose an interactive learning model where two tasks can not only interactively learn from each other but also generate predictions synchronously.

The main model structure is shown in Figure~\ref{fig:transformer}. First, the speech signal is processed into the acoustic feature sequence and projected by a linear transformation layer, whose dimension is converted to the hidden size $d_{\textrm{model}}$. Then, the encoder embeds the sequence into a high level acoustic representation. Two decoders are applied for different tasks in which one performs speech recognition and the other conducts speech translation. 

\begin{figure}[t]
	\centering
	\includegraphics[width=0.9\columnwidth]{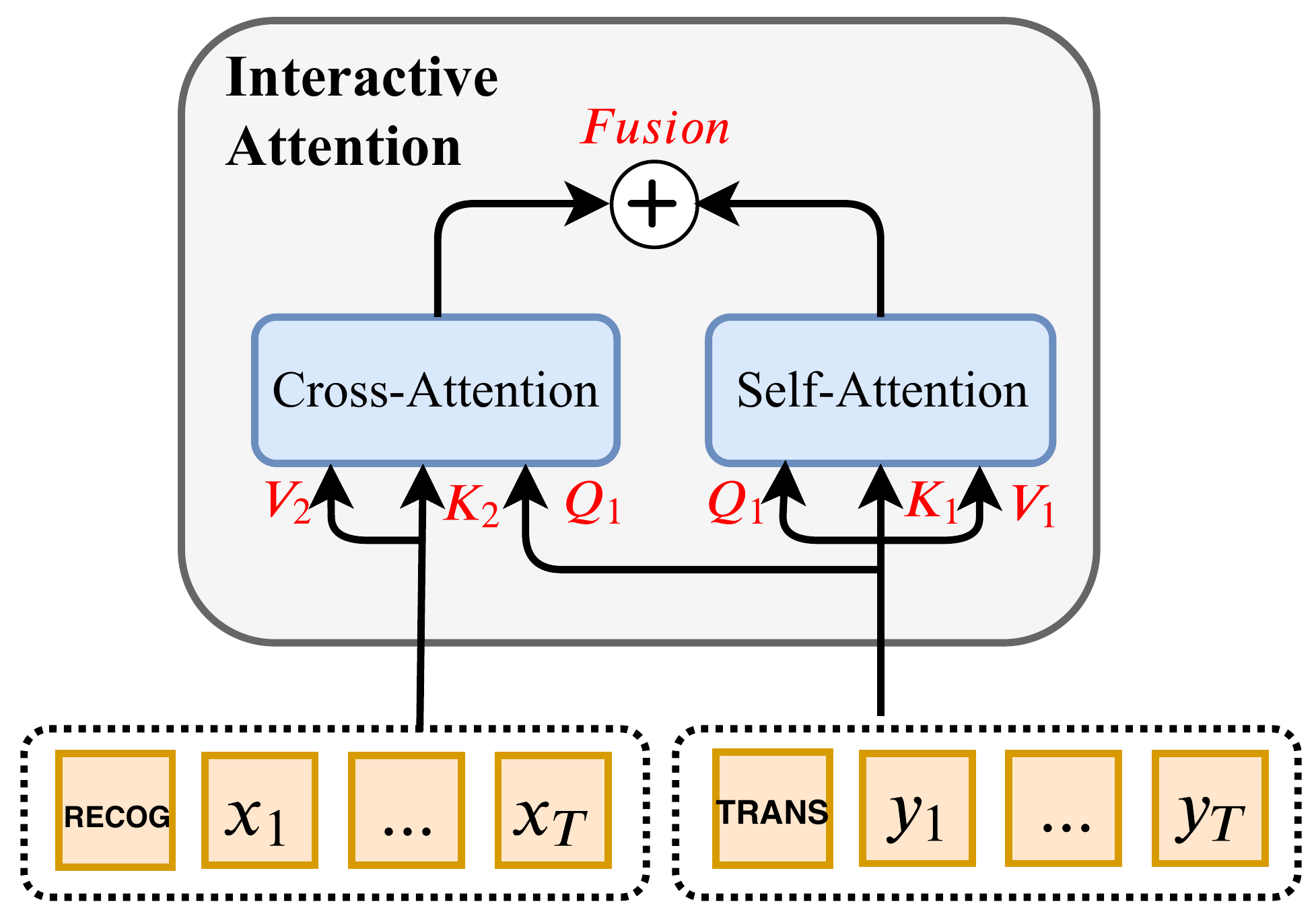}
	\caption{The interactive attention sub-layer consists of a self-attention sub-layer with a cross-attention sub-layer which can capture the information from the other task.}
	\label{fig:interactivate_learning}
\end{figure}

To make two decoders interactively learn from each other, we replace the self-attention sub-layer in the standard Transformer decoder with our proposed interactive attention sub-layer. 
As shown in Figure~\ref{fig:interactivate_learning}, the interactive attention sub-layer is composed of a self-attention sub-layer and a cross-attention sub-layer. The former uses the hidden representation from task 1 as the query $\mathbf{Q_1}$, key $\mathbf{K_1}$ and value $\mathbf{V_1}$ to learn higher representation $\mathbf{H_{self}}$. While the latter uses the hidden representation from task 1 as the query $\mathbf{Q_1}$, and the hidden representation from task 2 as key $\mathbf{K_2}$ and value $\mathbf{V_2}$ to integrate the representation $\mathbf{H_{cross}}$ of the other task. All the hidden representations are extracted from the same layer. It can be calculated as:
\begin{equation}
\mathbf{H_{self}}=\mathrm{Attention(\mathbf{Q_1},\mathbf{K_1},\mathbf{V_1})}
\end{equation}
\begin{equation}
\mathbf{H_{cross}}=\mathrm{Attention(\mathbf{Q_1},\mathbf{K_2},\mathbf{V_2})}
\end{equation}
Then the output of self-attention sub-layer and that of cross-attention sub-layer can be integrated by a fusion function to obtain the final representation:
\begin{equation}
\mathbf{H_{final}}=\mathrm{Fusion(\mathbf{H_{self}},\mathbf{H_{cross}})} 
\end{equation}
We use a linear interpolation as fusion function, which can be calculated as:
\begin{equation}
\mathbf{H_{final}}=\mathbf{H_{self}} + \lambda * \mathbf{H_{cross}}
\end{equation}
where $\lambda$ is a hyper-parameter to control how much information of the other task should be taken into consideration. 
Then both decoders can obtain the combined representation which contains information from the  outputs in two tasks. 

We apply interactive attention sub-layer to replace the self-attention sub-layer in the standard Transformer decoder, and it also utilizes the residual connections \cite{he2016deep} around each sub-layer, followed by layer normalization \cite{ba2016layer}. 
Other modules remain the same as standard Transformer model.

\subsection{Training and Inference}
\subsubsection{Training}
Since our approach performs ASR task and ST task in one model, two tasks can be optimized at the same time. We additionally append two special labels ($\langle  RECOG\rangle$ and $\langle TRANS\rangle$) at the start of transcriptions and translations to indicate whether the generation process is recognition or translation.  Given a set of training data $\small{D=\{\mathrm{S}^{(j)},\mathrm{X}^{(j)},\mathrm{Y}^{(j)}\}_{j=1}^{|D|}}$, where $\mathrm{S}$ is the sequence of speech features, $\mathrm{X}$ is the sequence of source transcription and $\mathrm{Y}$ is the corresponding target translation, the objective function is to maximize the log-likelihood over both the transcription and the translation,
\begin{equation}
\small
L(\theta)=\sum_{j=1}^{|D|} (\log P(\mathrm{X}^{(j)}|\mathrm{S}^{(j)},\mathrm{Y}^{(j)}) + \log P(\mathrm{Y}^{(j)}|\mathrm{S}^{(j)},\mathrm{X}^{(j)}))
\end{equation}

With interactive attention sub-layer, the recognition decoder and translation decoder can utilize the information from both itself and the other. Specifically, at time $i$, the recognition decoder and translation decoder have generated the first $i$$-$1 words respectively, then the $i$-th word in translation can be predicted based on the $i$$-$1 already generated translation words and the $i$$-$1 already transcribed words. It is the same for the generation process in speech recognition task. Therefore, the prediction probability of each transcription $ P(\mathrm{X}|\mathrm{S},\mathrm{Y})$ and translation $ P(\mathrm{Y}|\mathrm{S},\mathrm{X})$ can be formalized as,
\begin{align}
\log P(\mathrm{X}|\mathrm{S},\mathrm{Y}) &= \sum_{i=0}^{N-1} \log p(x_{i}|x_{<i},\mathrm{S},y_{<i}) \\
\log P(\mathrm{Y}|\mathrm{S},\mathrm{X}) &= \sum_{i=0}^{M-1} \log p(y_{i}|y_{<i},\mathrm{S},x_{<i})
\end{align} 

\subsubsection{Inference}
The inference process is similar with training. We run beam search algorithm for the two tasks. Two beams are applied for different tasks and expand hypotheses respectively. The outputs of two tasks are generated in parallel, with the interactive attention sub-layer to implement information exchanging between two decoders. At each step, the word with highest probability will be selected and added to each hypotheses. The inference process terminates until both tasks reach the end of sentences.  In this way, the hypotheses in speech recognition task and speech translation task can be generated synchronously. 

\subsection{Wait-$k$ Policy} 
Considering that the speech translation task is more difficult than speech recognition, it would be helpful for the translation process if the translation decoder can get more information at each step. 
Therefore, we introduce a wait-$k$ policy, in which the translation decoder begins to perform until the first $k$ source words are transcribed by the recognition decoder. That is, the generation process of translation is always $k$ words later than the generation of transcription. For example, if $k=2$, the first translation word is predicted based on the acoustic representation of encoder with the first two transcription words. Then the second translation word can use the hidden representation of acoustic encoder, the first three transcription words, and the first predicted translation word, etc.
~\citeauthor{ma2018stacl}~\shortcite{ma2018stacl}, they applied the wait-$k$ policy in simultaneous translation where the translation decoder is always k words behind the incoming source stream. Different from them, the decoders in our work have access to the complete source speech utterances, and the wait-$k$ policy is only applied to the translation decoder. During training, we append $k$ special label ($\langle DELAY\rangle$) before the start of translation, which indicates that the generation process of translation is $k$ steps later than recognition. 

\section{Experiments}
\subsection{Dataset}
Prior studies usually conduct experiments on Fisher and Callhome, a corpus of telephone conversations which include English transcriptions and Spanish (Es) translations \cite{post2013improved}. However, the ASR word error rate (WER) of this corpus is fairly high\footnote{This corpus contains ASR outputs which are provided by \citeauthor{post2013improved}~\shortcite{post2013improved}, with a WER of over 40\%.}, due to the spontaneous speaking style and challenging acoustics. 
Therefore, we construct a new speech translation corpus collected from TED talks which are a popular data resource in both speech recognition and machine translation fields.

To build this corpus, we first crawl the raw data (including video data, subtitles and timestamps) from the TED website\footnote{https://www.ted.com}. Audio in each talk is extracted from video and saved in \textit{wav} format. Subtitles in each talk usually have an English manual transcription and more than one translations in different languages. Here, we only collect the subtitles which contains English transcription with translations in German, French, Chinese, and Japanese (briefly, De/Fr/Zh/Ja). Adjacent subtitles and timestamps in English transcriptions are combined according to strong punctuations, such as period and question marks. Then each audio is segmented into small utterances based on the combined timestamps. This process guarantees that each speech utterance contains complete semantic information, which is important for translation. Translations in different languages are also combined based on the timestamps to align with speech utterances\footnote{\citeauthor{gangi2019must}~\shortcite{gangi2019must} built similar corpora, however their corpora do not consist of En-Zh and En-Ja language pairs, and they used a different segment way.}.

Finally, we obtain 235K/299K/299K/273K triplet data for En-De/Fr/Zh/Ja language pairs respectively, which contain speech utterances, manual transcriptions and translations. 
Development and test sets are split according to the partition in IWSLT. We use tst2014 as development (\textit{Dev})  set and tst2015 as \textit{test} set. The remaining data are used as training set. This dataset is available on http://www.nlpr.ia.ac.cn/cip/dataset.htm.

\begin{table*}[t]
	\centering
	\begin{tabular}{c|cc|cc|cc|cc}
		\hline
		\multirow{2}{*}{Model}    & \multicolumn{2}{c|}{En-De} & \multicolumn{2}{c|}{En-Fr} & \multicolumn{2}{c|}{En-Zh} & \multicolumn{2}{c}{En-Ja} \\
		\multicolumn{1}{c|}{}              & WER($\downarrow$)         & BLEU($\uparrow$)         & WER($\downarrow$)          & BLEU($\uparrow$)         & WER($\downarrow$)         & BLEU($\uparrow$)         & WER($\downarrow$)          & BLEU($\uparrow$)         \\ \cline{1-9}
		Text MT              & /              & 22.19           & /               & 30.68           & /              & 25.01           & /               & 22.93           \\
		\cline{1-9}
		Pipeline                                   & 16.19          & 19.50           & 14.20            & 26.62           & 14.20          & 21.52            & 14.21            &  \textbf{20.87}           \\
		E2E                                & 16.19          & 16.07           & 14.20            & 27.63           & 14.20          & 19.15           & 14.21            & 16.59           \\
		Multi-task                               & 15.20          & 18.08           & 13.04           & 28.71           & 13.43          & 20.60            & 14.01           & 18.73           \\
		Two-stage                              & 15.18          & 19.08           & 13.34           & \textbf{30.08}           & 13.55          & 20.99           & 14.12           & 19.32           \\
		\cline{1-9}
		Interactive   & \textbf{14.76}          & \textbf{19.82$_{\star}^{\ast}{\ddagger}$}           & \textbf{12.87$_{\star}^{\ast}{\ddagger}$}          & 29.79$_{\star}^{\ast}$           &  \textbf{13.38$^{\ast}$}          &  \textbf{21.68$_{\star}{\ddagger}$}           &  \textbf{13.91$_{\star}^{\ast}{\ddagger}$}           & 19.60$_{\star}\ddagger$           \\ 
		\hline
	\end{tabular}
	\caption{Evaluation of speech recognition and speech translation on TED En-De/Fr/Zh/Ja datasets. E2E denotes to the pretrained end-to-end ST model, and Interactive represents our proposed interactive learning model. 
		$\ast$, $\star$, and $\ddagger$ indicate Interactive learning model is statistically significant ($p<0.01$) compared with Pipeline, Multi-task, and Two-stage, respectively.}
	\label{tbl:overall}
\end{table*}

\subsection{Model Settings}
The speech features have 80-dimension log-Mel filterbanks extracted with a step size of 10ms and window size of 25ms, which are extended with mean subtraction and variance normalization. The features are stacked with 3 frames to the left and downsampled to a 30ms frame rate. 
We remove punctuations, lowercase and tokenize  English transcriptions using scripts from Moses\footnote{https://www.statmt.org/moses/}. 
We also lowercase and tokenize the translations in German and French.
Chinese sentences are segmented by Jieba\footnote{https://github.com/fxsjy/jieba} and Japanese sentences are segmented by Mecab\footnote{http://taku910.github.io/mecab}. 
For En-De and En-Fr, parallel sentences are encoded using BPE method \cite{sennrich2015neural} which has a shared vocabulary of 30K tokens. For En-Zh and En-Ja, we encode source transcriptions and target translations, respectively, and the vocabulary size is limited to the most frequent 30K. 
ASR performance is evaluated with WER computed on lowercased, tokenized manual transcriptions without punctuations. As for text translation and speech translation, we report case-insensitive tokenized BLEU \cite{papineni2002bleu} for De/Fr language pairs and character-level BLEU for Zh/Ja.

All of the models are implemented based on the model adopted from Transformer. We use the configuration \textit{transformer\_base} used by~\citeauthor{vaswani2017attention}~\shortcite{vaswani2017attention} which contains 6-layer encoders and 6-layer decoders with 512-dimensional hidden sizes. 
We train our models with Adam optimizer \cite{kingma2014adam} on 2 NVIDIA V100 GPUs. For inference, we perform beam search with a beam size of $4$.

\subsection{Baselines}
We compare the proposed method with the following baseline models:
\begin{itemize}
	\item Pipeline system: ASR and MT model are independently trained, and then the outputs of ASR model are taken as the inputs to MT model.
	\item Pretrained ST model: The encoder of end-to-end ST model is first initialized by training on ASR data, and then the model is finetuned on speech translation data.
	\item Multi-task learning model: ASR model and ST model are jointly trained with the parameters of encoder shared.
	\item Two-stage model: This model contains two stages where the outputs of the first stage are transcriptions and the second stage are translations. We re-implement the basic model based on Transformer following~\citeauthor{sperber2019attention}~\shortcite{sperber2019attention}. The model in the first stage is also initialized by training on ASR data.
\end{itemize}

\subsection{Results}
Table \ref{tbl:overall} shows the main results of speech recognition and speech translation on En-De/Fr/Zh/Ja TED corpora. The BLEU scores in the first row are the translation results by text MT model when the clean manual transcriptions are given as inputs. This can be seen as the upper bound for speech translation task. 
We set $\lambda=0.3$ and $k=3$ in the interactive learning model. 

\subsubsection{Similar Languages}
We first analyze En-De and En-Fr language pairs. From the first two rows, we can see that the translation quality drops dramatically when the output of ASR model is fed as the input to the MT model compared with the clean transcriptions input. It indicates that text MT model is very sensitive to recognition errors, which is one of the main problems in the pipeline system.
Pretrained end-to-end  ST model outperforms the pipeline system by 0.99 BLEU points on En-Fr language direction, but it does not show superiority on En-De. 
We argue that end-to-end model may have superiority of less error propagation on more similar language pairs, such as En-Fr or En-Es. This is consistent with \citeauthor{weiss2017sequence}~\shortcite{weiss2017sequence} who conducted experiments on En-Es and found end-to-end ST has better performance than the pipeline system. 
Compared with the end-to-end model, multi-task learning model can obtain some improvements, which improves 2.01 and 0.98 BLEU scores for En-De and En-Fr, respectively.
However, with information exchanging, our proposed interactive learning model significantly outperforms multi-task learning model on the quality of both speech recognition and speech translation. 
It demonstrates the effectiveness of the interactive attention mechanism. 
Although our method does not outperform two-stage model on En-Fr speech translation task, it has a better performance on ASR result. The underlying reason is that the goal of two-stage model is to optimize the translation quality with the information of complete transcription while ignoring the recognition, so it can improve the translation quality but leave the recognition alone. 

\subsubsection{Dissimilar Languages}
It is even more difficult to implement end-to-end speech translation on dissimilar language pairs, such as En-Zh and En-Ja. Because these kind of models are required to learn not only the alignments between source frames and translation words, but also the word orders in long distances. Therefore, in our experiments, most of the end-to-end models are inferior than pipeline system. However, the proposed interactive learning model can significantly outperform end-to-end ST model, traditional multi-task learning model and two-stage model, approaching to or slightly better than  pipeline system.

\subsection{Effect of the Hyper-parameters}
\label{sec:lambda}
\begin{table}[t]
	\centering
	\begin{tabular}{@{}c|cc|cc@{}}
		\hline
		\multirow{2}{*}{$\lambda$} & \multicolumn{2}{c|}{Dev} & \multicolumn{2}{c}{Test} \\  
		\multicolumn{1}{c|}{}    & WER      & BLEU     & WER      & BLEU     \\ 
		\cline{1-5}
		0.0                                       & 14.87       & 15.74       & 13.43       & 20.60       \\
		0.1                                        & \textbf{14.47 }      & 15.93       & \textbf{12.92  }     & 20.88       \\
		0.3                                       &  14.51     &    \textbf{16.28 }     & 13.24       &      \textbf{21.01}  \\
		0.5                                       &  15.50        & 15.66       & 14.17       & 20.68        \\ 
		1.0                                       &   15.92         &  15.06          &     14.52        &    20.13        \\ 
		\hline
	\end{tabular}
	\caption{The performance of speech recognition and speech translation under different hyper-parameters $\lambda$ on the En-Zh \textit{Dev} set and \textit{Test} set.}
	\label{tbl:hyper-parameter}
\end{table}
We investigate how much information from two tasks should be taken into consideration in the interactive attention sub-layer. Table \ref{tbl:hyper-parameter} reports the WER and BLEU scores under different $\lambda$ on the En-Zh. If $\lambda=0.0$, the model degrades to traditional multi-task learning model which does not utilize any information from the other task. As shown in the table, as $\lambda$ increases, both recognition quality and translation quality can be improved with information interacting. When $\lambda=0.3$, our interactive learning model achieves the best performance on the speech translation task.  However, $\lambda$ can not be too large, otherwise two tasks may  interfere with each other and affect its own performance. Therefore, we use $\lambda=0.3$ for all experiments.

\subsection{Effect of $k$ in Wait-$k$ Policy}
\label{sec:k}
\begin{table}[t]
	\centering
	\begin{tabular}{@{}l|cc|cc@{}}
		\hline
		\multirow{2}{*}{Wait-$k$} & \multicolumn{2}{c|}{Dev} & \multicolumn{2}{c}{Test} \\ 
		\multicolumn{1}{c|}{}              & WER      & BLEU     & WER      & BLEU     \\ 
		\hline
		Wait-0                                        & 14.51        & 16.28       & 13.24       & 21.01       \\
		Wait-1                                         & 14.29       & 16.09       & \textbf{13.17}       & 21.30       \\
		Wait-3                                        & \textbf{14.24}      &  \textbf{16.74}       & 13.38       & \textbf{21.68}       \\
		Wait-5                                        & 14.36       &  16.55      & 13.51       & 21.45     \\  
		\hline
	\end{tabular}
	\caption{The performance of speech recognition and speech translation with different word latency in wait-$k$ policy on the En-Zh \textit{Dev} set and \textit{Test} set.}
	\label{tbl:wait-k}
\end{table}

We then investigate the effect of word latency in wait-$k$ policy on En-Zh language pairs. As shown in Table \ref{tbl:wait-k}, the speech translation quality in BLEU scores can be improved with the increase of word latency. It indicates that the speech translation task can become easier if more source information from the same modality is given. However, as $k$ increases, it will affect the performance of speech recognition task. 
If $k \rightarrow \infty$,  this model degrades to the analogous two-stage model. Then the speech translation task can obtain the information from complete transcribed sentence, while speech recognition task can not utilize any information from translations. 
The interactive learning model has the best performance when $k=3$.  

\subsection{Parameters and Speeds}
\begin{table}[t]
	\centering
	\begin{tabular}{l|c|c|c}
		\hline
		\multirow{2}{*}{Model} & \multirow{2}{*}{Params} & \multicolumn{2}{c}{Speed} \\ 
		\cline{3-4} 
		~                          &            ~            & Train       & Inference        \\ 
		\hline
		Pipeline               & 122.4M              & /               & 10.89       \\
		E2E           & 61.2M                & 4.73         & 16.17       \\
		Multi-task            & 61.2M                & 4.41         & 16.26       \\
		Two-stage          & 92.7M                & 1.13         & 7.44        \\
		Interactive           & 61.2M                & 4.23        & 11.98       \\ \hline
	\end{tabular}
	\caption{Statistics of parameters, training and inference speeds. The number in Train denotes the average number of training steps per second. The number in Inference is the average amount of sentences generated per second.}
	\label{tbl:parameter}
\end{table}
The parameter sizes of different models are shown in Table \ref{tbl:parameter}. The pipeline system needs a separate ASR model and MT model, so its parameters are doubled. 
Two-stage model has 1.5 times larger parameters since it has two different decoders in two stages. 
In multi-task learning model and interactive learning model, we share the parameter between different tasks. Therefore, they have the same number of parameters with end-to-end model. 
Table \ref{tbl:parameter} also shows the training and inference speed of different models on En-Zh  test set. The training speed of interactive learning model is 4.23 steps per second, which is comparable with the end-to-end model but is much faster than two-stage model. During inference, the average decoding speed of interactive learning model is 11.98 utterances per second. Although it is slower than end-to-end model and multi-task learning model, it can generate transcriptions paired with translations in one model synchronously. While two-stage model can also generate transcription and translation in a single model, its implementation which is in a cascade manner is much slower even than pipeline system.

\subsection{Case Study}
\begin{figure}[t]
	\centering
	\includegraphics[width=0.95\columnwidth]{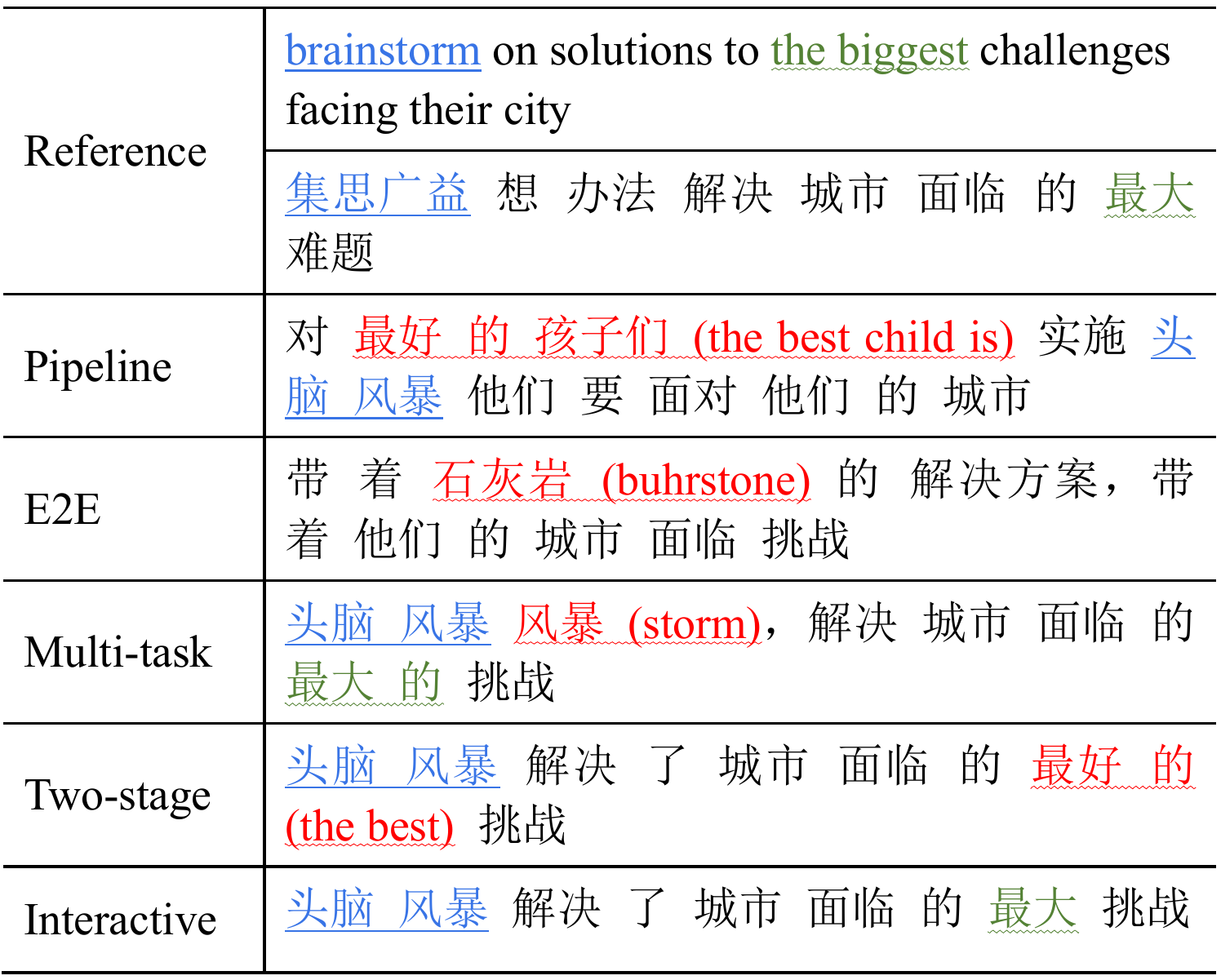}
	\caption{An Example of  speech translation generated by different models. Words in blue and green are original words in the manual transcription, corresponding translation reference and correct translations with the similar meaning, while words in red are the wrong translations.}
	\label{fig:case}
\end{figure}

We show the case study in Figure~\ref{fig:case}. 
In pipeline system,  ASR model first recognizes the speech utterance into ``brainstormed on solutions to the best child is facing their city". Since it wrongly recognizes ``the biggest challenges" into ``the best child is", text MT then translates the incorrect recognition phrase, resulting the result is far from the reference. It is more difficult for the end-to-end ST model to generate a correct translation and its output is totally wrong. This model may comprehend the speech of ``brainstorm" into ``buhrstone" which has a similar pronunciation and it omits the translation of ``the biggest". Although the multi-task learning model has an enhanced acoustic encoder, it repeatedly attends to the speech of ``storm" without  transcription as guidance and translates it twice. As for two-stage model, it erroneously recognized ``the biggest" into ``the best" in the first stage based on which the second decoder also gives a wrong translation. Compared to the above approaches, our model generates the right transcription and translation through interactive attention mechanism, which matches the reference best.

\section{Conclusion and Future Work}
In this paper, we propose an interactive learning model to conduct speech recognition and speech translation interactively and simultaneously. The generation process of recognition and translation in this model can not only utilize the already generated outputs, but also the outputs generated in the other task. We then present a wait-$k$ policy which can further improve the speech translation quality. Experimental results on different language pairs demonstrate the effectiveness of our model. In the future, we plan to design a streaming encoder and make a step forward in achieving end-to-end simultaneous interpretation. 

\section{ Acknowledgments}
The research work described in this paper has been supported by the National Key Research and Development Program of China under Grant No. 2016QY02D0303, the Natural Science Foundation of China under Grant No. U1836221 and 61673380, and Beijing Municipal Science and Technology Project No. Z181100008918017 as well. The research work in this paper has also been supported by Beijing Advanced Innovation Center for Language Resources.

\bibliographystyle{aaai} 
\bibliography{reference}

\end{document}